\documentclass{article}

\usepackage{microtype}
\usepackage{graphicx}
\usepackage{subfigure}
\usepackage{booktabs} 
\usepackage{listings} 
 
\usepackage{hyperref}

\usepackage[accepted]{icml2019}

\usepackage{amsmath,amsfonts,bm}

\def\eqref#1{equation~\ref{#1}}

\def\1{\bm{1}}

\def\rx{{\textnormal{x}}}
\def\ry{{\textnormal{y}}}
\def\rz{{\textnormal{z}}}

\DeclareMathAlphabet{\mathsfit}{\encodingdefault}{\sfdefault}{m}{sl}
\SetMathAlphabet{\mathsfit}{bold}{\encodingdefault}{\sfdefault}{bx}{n}

\newcommand{\E}{\mathbb{E}}

\newcommand{\KL}{D_{\mathrm{KL}}}
\newcommand{\Var}{\mathrm{Var}}

\newcommand{\normdiff}{\left\Vert d \right\Vert }

\icmltitlerunning{WAIC, but Why? Generative Ensembles for Robust Anomaly Detection}

\begin{document}

\twocolumn[
\icmltitle{WAIC, but Why? \\ Generative Ensembles for Robust Anomaly Detection}



\icmlsetsymbol{equal}{*}

\begin{icmlauthorlist}
\icmlauthor{Hyunsun Choi}{equal}
\icmlauthor{Eric Jang}{equal,goo}
\icmlauthor{Alexander A. Alemi}{goo}
\end{icmlauthorlist}

\icmlaffiliation{goo}{Google Inc.}

\icmlcorrespondingauthor{Hyunsun Choi}{hunsun1005@gmail.com}
\icmlcorrespondingauthor{Eric Jang}{ejang@google.com}
\icmlcorrespondingauthor{Alex Alemi}{alemi@google.com}

\icmlkeywords{Machine Learning, ICML}

\vskip 0.3in
]



\printAffiliationsAndNotice{\icmlEqualContribution} 

\begin{abstract}
Machine learning models encounter Out-of-Distribution (OoD) errors when the data seen at test time are generated from a different stochastic generator than the one used to generate the training data. One proposal to scale OoD detection to high-dimensional data is to learn a tractable likelihood approximation of the training distribution, and use it to reject unlikely inputs. However, likelihood models on natural data are themselves susceptible to OoD errors, and even assign large likelihoods to samples from other datasets. To mitigate this problem, we propose Generative Ensembles, which robustify density-based OoD detection by way of estimating epistemic uncertainty of the likelihood model. We present a puzzling observation in need of an explanation -- although likelihood measures cannot account for the typical set of a distribution, and therefore should not be suitable on their own for OoD detection, WAIC performs surprisingly well in practice.
\end{abstract}


\section{Introduction}
\label{introduction}

Knowing when a machine learning (ML) model is qualified to make predictions on an input is critical to safe deployment of ML technology in the real world. When training and test distributions differ, neural networks may provide -- with high confidence -- arbitrary predictions on inputs that they are unaccustomed to seeing. This is known as the Out-of-Distribution (OoD) problem. In addition to ML safety, identifying OoD inputs (also referred to as anomaly detection) is a crucial feature of many data-driven applications, such as credit card fraud detection and monitoring patient health in medical settings.

A typical OoD scenario is as follows: a machine learning model infers a predictive distribution $p(\ry|\rx)$ from input $\rx$, which at training time is sampled from a distribution $p(\rx)$. At test-time, $p(\ry|\rx)$ is evaluated on a single input sampled from $q(x)$, and the objective of OoD detection is to infer whether $p(\rx) \equiv q(\rx)$. This problem may seem ill-posed at first glance, because we wish to compare $p(\rx)$ and $q(\rx)$ using only a single sample from $q(x)$. Nevertheless, this setup is common when serving ML model predictions over the Internet to anonymous, untrusted users, and is also assumed in adversarial ML literature. Each user generates data from a unique test set $q(x)$ and may only request one prediction. 

One approach to OoD detection is to combine a dataset of anomalies with in-distribution data and train a binary classifier to tell them apart, or alternatively, appending a ``None of the above'' category to a classification model. The classifier then learns a decision boundary, or likelihood ratio, between $p(\rx)$ and the anomaly distribution $q(\rx)$. However, the discriminative approach to anomaly detection requires $q(\rx)$ to be specified at training time; this is a severe flaw when anomalous data is rare (e.g. medical seizures) or not known ahead of time, e.g. generated by an adversary. 

On the other hand, density estimation techniques do not assume an anomaly distribution at training time, and can be used to assign lower likelihoods to OoD inputs \cite{bishop1994novelty}. However, we present a couple concerns about the appropriateness of likelihood models for OoD detection.

\subsection{Counterintuitive Properties of Likelihood Models}

When $p(x)$ is unknown, a generative model $p_\theta(\rx)$, parameterized by $\theta$, can be trained to approximate $p(\rx)$ from its samples. Generative modeling algorithms have improved dramatically in recent years, and are capable of learning probabilistic models over massive, high-dimensional datasets such as images, video, and natural language \citep{kingma2018glow, vaswani2017attention, wang2018high}. Autoregressive Models and Normalizing Flows (NF) are fully-observed likelihood models that construct a tractable log-likelihood approximation to the data-generating density $p(\rx)$ \citep{uria2016neural, dinh2014nice, rezende2015variational}. Variational Autoencoders (VAE) are latent variable models that maximize a variational lower bound on log density \citep{kingma2013auto, rezende2014stochastic}. Finally, Generative Adversarial Networks (GAN) are implicit density models that minimize a divergence metric between $p(\rx)$ and generative distribution $p_\theta(\rx)$ \citep{goodfellow2014generative}.


Likelihood models implemented using neural networks are susceptible to malformed inputs that exploit idiosyncratic computation within the model \citep{szegedy2013intriguing}. When judging natural images, we assume an OoD input $x \sim q(\rx)$ should remain OoD within some $L^P$-norm, and yet a Fast Gradient Sign Method (FGSM) attack \citep{goodfellow6572explaining} on the predictive distribution can realize extremely high likelihood predictions \citep{nguyen2015deep}. Conversely, a FGSM attack in the reverse direction on an in-distribution sample $x \sim p(\rx)$ creates a perceptually identical input with low likelihood \citep{kos2018adversarial}.

\begin{figure}[h]
  \centering
  \includegraphics[width=\linewidth]{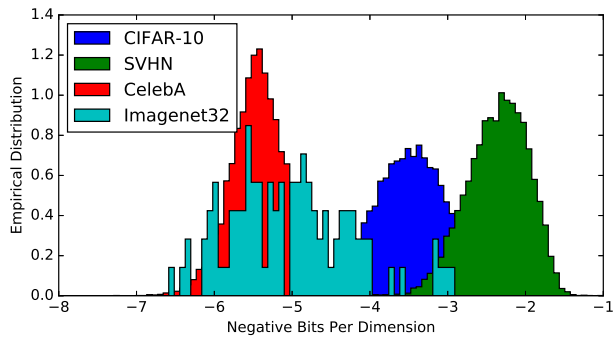}
  \caption{Density estimation models are not robust to OoD inputs. A GLOW model \citep{kingma2018glow} trained on CIFAR-10 assigns much higher likelihoods to samples from SVHN than samples from CIFAR-10. .}
  \label{fig:single_model_failure}
\end{figure}

An earlier version of this paper, concurrently with work by Nalisnick et al. \yrcite{nalisnick2018dodeep}, and Hendrycks et al. \yrcite{hendrycks2018deep} showed that likelihood models can be fooled by OoD datasets that are not even adversarial by construction. As shown in Figure~\ref{fig:single_model_failure}, a likelihood model trained on the CIFAR-10 dataset yields higher likelihood predictions on SVHN images than the CIFAR-10 training data itself! For a flow-based model with an isotropic Gaussian prior, this implies that SVHN images are systematically projected closer to the origin than the training data \cite{nalisnick2018dodeep}.

\subsection{Likelihood and Typicality}
\label{sec:typicality}
Even if we could compute likelihoods exactly, it may not be a sufficient measure for scoring OoD inputs. It is tempting to suggest a simple one-tailed test in which lower likelihoods are OoD, but the intuition that in-distribution inputs should have the highest likelihoods \emph{does not hold in higher dimensions}. For instance, consider an isotropic 784-dimensional Gaussian. A data point at the origin has maximum likelihood, and yet it is highly atypical because the vast majority of probability mass lies in an annulus of radius $\sqrt{784}$. Likelihoods can determine whether a point lies in the support of a distribution, but do not reveal where the probability mass is concentrated. 

The main contributions of this work are as follows. First, we observe that likelihood models assign higher densities to OoD datasets than the ones they are trained on (SVHN for CIFAR-10 models, and MNIST for Fashion MNIST models). Second, we propose Generative Ensembles, an anomaly detection algorithm that combines density estimation with uncertainty estimation. Generative Ensembles are trained independently of the task model, and can be implemented using exact or approximate likelihood models. We also demonstrate how predictive uncertainty can be applied to robustify implicit GANs and leverage them for anomaly detection. Finally, we present yet another surprising property of deep generative models that warrants further explanation: density estimation should not be able to account for probability mass, and yet Generative Ensembles outperform OoD baselines on the majority of common OoD detection problems, and demonstrate competitive results with discriminative classification approaches on the Kaggle Credit Fraud dataset.

\section{Related Work}
\label{related_work}

Anomaly detection methods are closely intertwined with techniques used in uncertainty estimation, adversarial defense literature, and novelty detection.

\subsection{Uncertainty Estimation}

OoD detection is closely related to the problem of uncertainty estimation, whose goal is to yield calibrated confidence measures for a model's predictive distribution $p(\ry|\rx)$. Well-calibrated uncertainty estimation integrates several forms of uncertainty into $p(\ry|\rx)$: model mispecification uncertainty (OoD detection of invalid inputs), aleatoric uncertainty (irreducible input noise for valid inputs), and epistemic uncertainty (unknown model parameters for valid inputs). In this paper, we study OoD detection in isolation; instead of considering whether $p(\ry|\rx)$ should be trusted for a given $x$, we are trying to determine whether $x$ should be fed into $p(\ry|\rx)$ at all. 

Predictive uncertainty estimation is a model-dependent OoD technique because it depends on task-specific information (such as labels and task model architecture) in order to yield an integrated estimate of uncertainty. ODIN \citep{liang2017enhancing}, MC Dropout \citep{gal2016dropout} and DeepEnsemble \citep{lakshminarayanan2017simple} model a calibrated predictive distribution for a classification task. Variational information bottleneck (VIB) \citep{alemi2018uncertainty} performs divergence estimation in latent space to detect OoD, but is technically a model-dependent technique because the latent code is trained jointly with the downstream classification task. 
 
One limitation of model-dependent OoD techniques is that they may discard information about $p(\rx)$ in learning the task-specific model $p(y|x)$. Consider a contrived binary classification model on images that learns to solve the task perfectly by discarding all information except the contents of the first pixel (no other information is preserved in the features). Subsequently, the model yields confident predictions on any distribution that happens to preserve identical first-pixel statistics. In contrast, density estimation in data space $x$ considers the structure of the entire input manifold, without bias towards a particular downstream task or task-specific compression. 

In our work we estimate predictive uncertainty of the scoring model itself. Unlike predictive uncertainty methods applied to the task model's predictions, Generative Ensembles do not require task-specific labels to train. Furthermore, model-independent OoD detection aids interpretation of predictive uncertainty by isolating the uncertainty component arising from OoD inputs.

\subsection{Adversarial Defense}

Although adversarial attack and defense literature usually considers small $L^p$-norm modifications to input (demonstrating the alarming sensitivity of neural networks), there is no such restriction in practice to the degree with which an input can be perturbed in a test setting. We consider adversarial inputs in the broader context of the OoD problem, where inputs can be swapped with other datasets, transformed and corrupted \cite{hendrycks2018benchmarking}, or are explicitly designed to fool the model.

Song et al. \yrcite{song2017pixeldefend} observe that adversarial examples designed to fool a downstream task tend to have low likelihoods under an independent generative model. They propose a ``data purification'' pipeline, where inputs are modified via gradient ascent on model likelihood before being passing to the classifier. Their evaluations are restricted to $L^p$-norm attacks on in-distribution inputs on the classifier, and do not take into account that the generative model itself may be susceptible to OoD errors. In fact, gradient ascent on model likelihood has the exact opposite of the desired effect when the input is OoD to begin with. In our experiments we measure the degree to which we can identify adversarially perturbed ``rubbish inputs'' as anomalies, and also note that adversarial examples for IWAE predictions have high rates under the latent code, making them suitable for anomaly detection.

\subsection{Novelty Detection}

When learning signals are scarce, such as in reinforcement learning (RL) with sparse rewards, the anomaly detection problem is re-framed as \emph{novelty detection}, whereby an agent attempts to visit states that are OoD with respect to previous experience \citep{fu2017ex2, marsland2003novelty}. Anomaly detection algorithms, including our proposed Generative Ensembles, are directly applicable as novelty bonuses for exploration. However, we point out in Section~\ref{sec:typicality} that likelihoods are deceiving in high dimensions - the point of highest density under a high-dimensional state distribution may be exceedingly rare.

\section{Generative Ensembles}

We introduce Generative Ensembles, a novel anomaly detection algorithm that combines likelihood models with predictive uncertainty estimation via ensemble variance. Concretely, an ensemble of generative models that compute exact or approximate likelihoods (autoregressive models, flow-based models, VAEs) are used to estimate the Watanabe-Akaike Information Criterion (WAIC).

\subsection{Watanabe Akaike Information Criterion (WAIC)}
First introduced in \citet{watanabe2010asymptotic}, 
the Watanabe-Akaike Information Criterion (WAIC) gives an 
asymptotically correct estimate of the gap between the 
training set and test set expectations.  
If we had access to samples from the true 
Bayesian posterior of a model, we could 
compute a corrected version of the expected log 
~\citep{watanabegrey}\footnote{
    Note that in this work we form a robust measure of what Watanabe calls the \emph{Gibbs} loss ($\E_{\theta} [ \log p_\theta (\rx) ]$),
    which while distinct from the \emph{Bayes} loss ($\log \E_{\theta}[ p_\theta(\rx)]$) has the same gap asymptotically~\citep{watanabegrey}.
}:
\begin{equation}
\text{WAIC}(x) = \E_\theta[\log p_\theta(\rx)] - \Var_\theta\left[\log p_\theta(\rx)\right]
\end{equation}
The correction term subtracts the variance in likelihoods 
across independent samples from the posterior.
This acts to robustify our estimate, ensuring that points
that are sensitive to the particular choice of posterior parameters
are penalized.

In this work we do not have exact posterior samples,
so we instead utilize independently trained model instances as a 
proxy for posterior samples, following
\cite{lakshminarayanan2017simple}.
Being trained with Stochastic Gradient Descent (SGD),
the independent models in the ensemble act as approximate 
posterior samples~\citep{mandt2017stochastic}.

\subsection{Does WAIC Address Typicality?}

\begin{figure}[h]
  \centering
  \includegraphics[width=\linewidth]{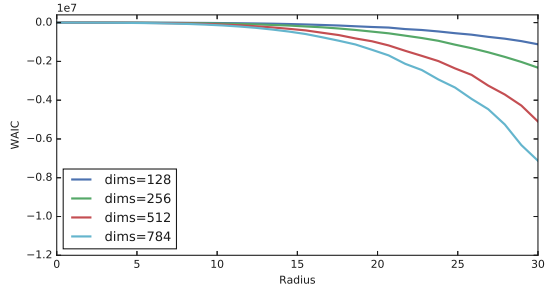}
  \caption{WAIC estimated using Jackknife resampling of data points drawn from an isotropic Gaussian, for an ensemble of size $N=10$. Lines correspond to a dimensionality of the data.}
  \label{fig:waic-decreases}
\end{figure}

Although WAIC can protect models against likelihood estimation errors, we show via a toy model that it should not be able to distinguish whether a point is in the typical set. We illustrate this phenomena in Figure~\ref{fig:waic-decreases} on ensembles of Gaussians fitted to samples from an isotropic Gaussian with leave-one-out cross validation. Like likelihoods, the WAIC measure also decreases monotonically from the origin.

\begin{figure}[h]
  \centering
  \includegraphics[width=\linewidth]{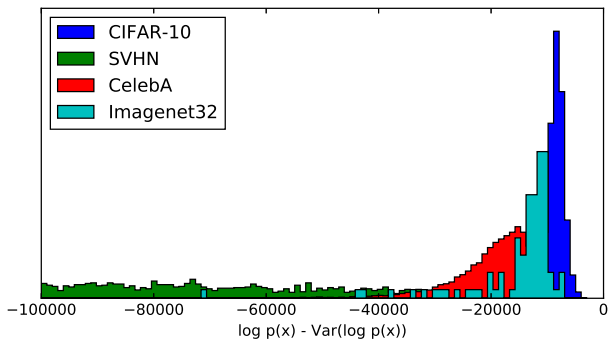}
  \caption{We use ensembles of generative models to implement the Watanabe-Akaike Information Criterion (WAIC), which combines density estimation with uncertainty estimation. Histograms correspond to predictions over test sets from each dataset.}
  \label{fig:waic_fix}
\end{figure}

Given that SVHN latent codes lie interior to the CIFAR-10 annulus (Figure~\ref{fig:single_model_failure}), WAIC should fail to distinguish SVHN as OoD. To our surprise, we show in Figure~\ref{fig:waic_fix} and Table~\ref{table:results_table} that not only does WAIC reject SVHN as OoD, but it outperforms a baseline that does test for typicality by measuring the Euclidean distance to the origin!

\begin{figure}[h]
  \centering
  \includegraphics[width=\linewidth]{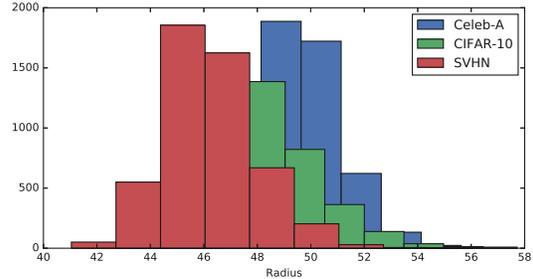}
  \caption{Histogram of euclidean norms in latent space. Glow models map CIFAR-10, SVHN, and Celeb-A outside of the typical set.}
  \label{fig:radii_interior}
\end{figure}

Glow models ought to map the training distribution (CIFAR-10) to a distribution whose probability mass is concentrated in an annulus of radius $\sqrt{3072} (\approx 55.4)$, but as we show in Figure~\ref{fig:radii_interior}, the Glow model actually maps CIFAR, SVHN, and Celeb-A to annuli whose mean radii span the range $42-54$. In our experiments, the variance of radii across models is larger for SVHN and Celeb-A than CIFAR-10, allowing the WAIC metric to correctly identify these as anomalies despite prior hypotheses that WAIC is insufficient.

\subsection{Generative Ensembles of GANs}
\label{sec:gans}

We describe how to improve GAN-based anomaly detection with ensembles. Although we cannot readily estimate WAIC with GANs, we can still leverage the principle of epistemic uncertainty to improve a GAN discriminator's ability to detect OoD inputs.

\begin{figure}[h]
  \centering
  \subfigure[]{\includegraphics[width=.31\linewidth]{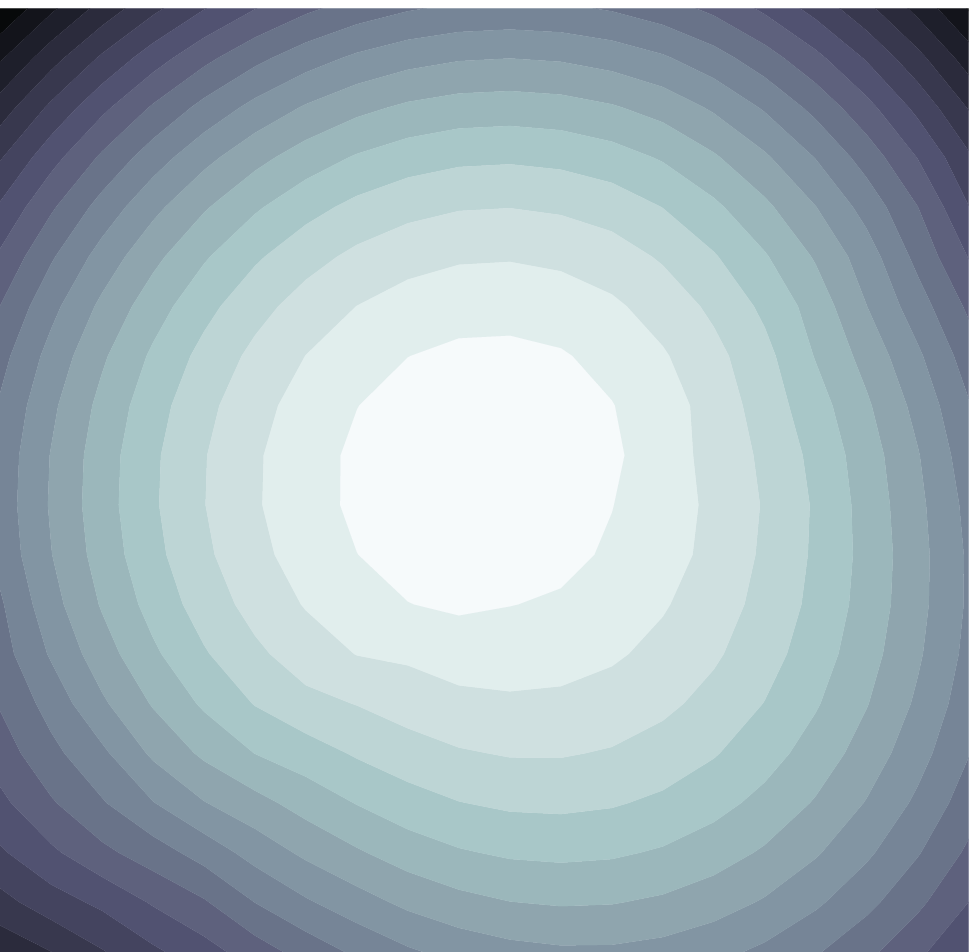}} \hspace{1pt}
  \subfigure[]{\includegraphics[width=.31\linewidth]{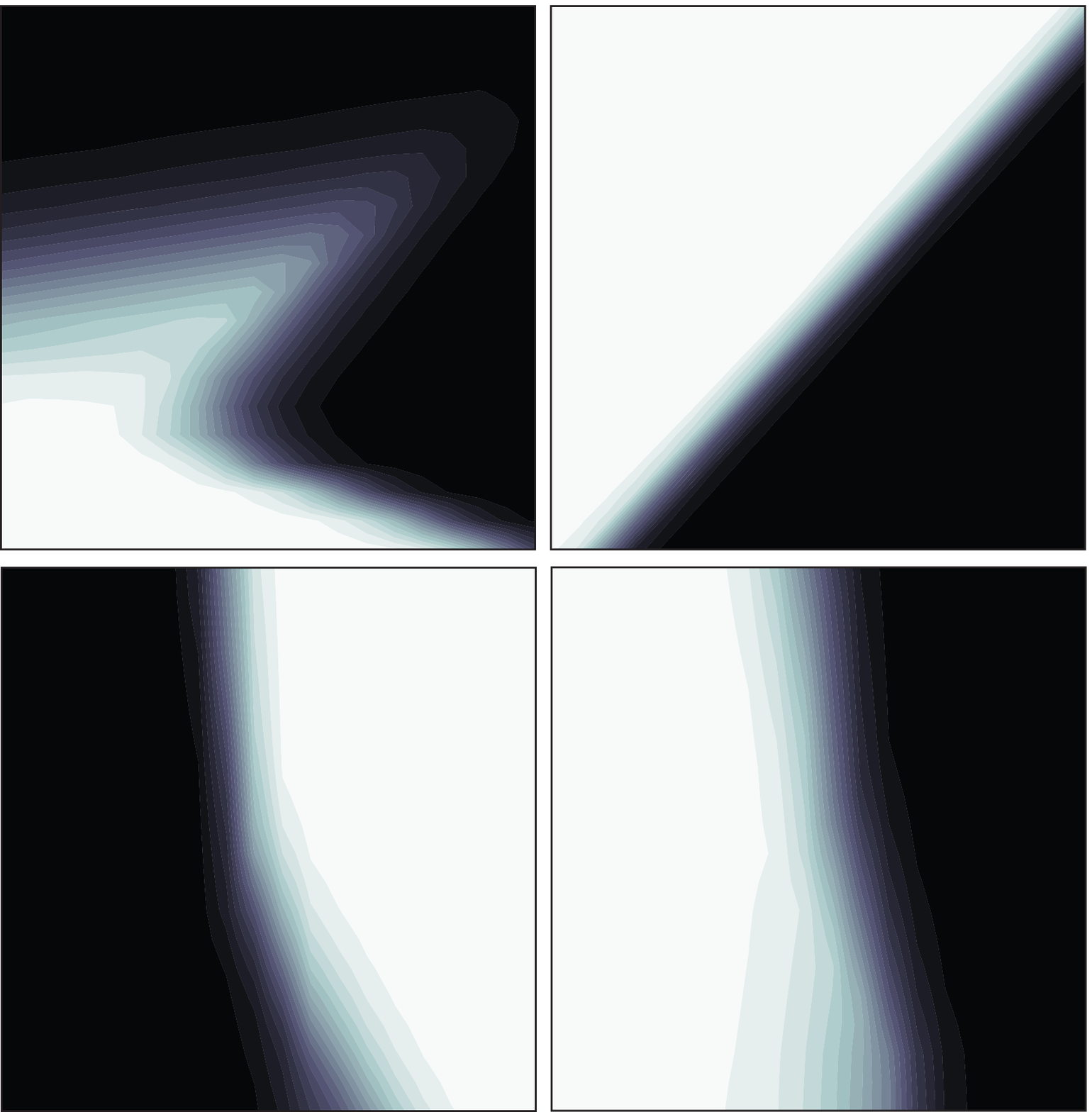}} \hspace{1pt}
  \subfigure[]{\includegraphics[width=.31\linewidth]{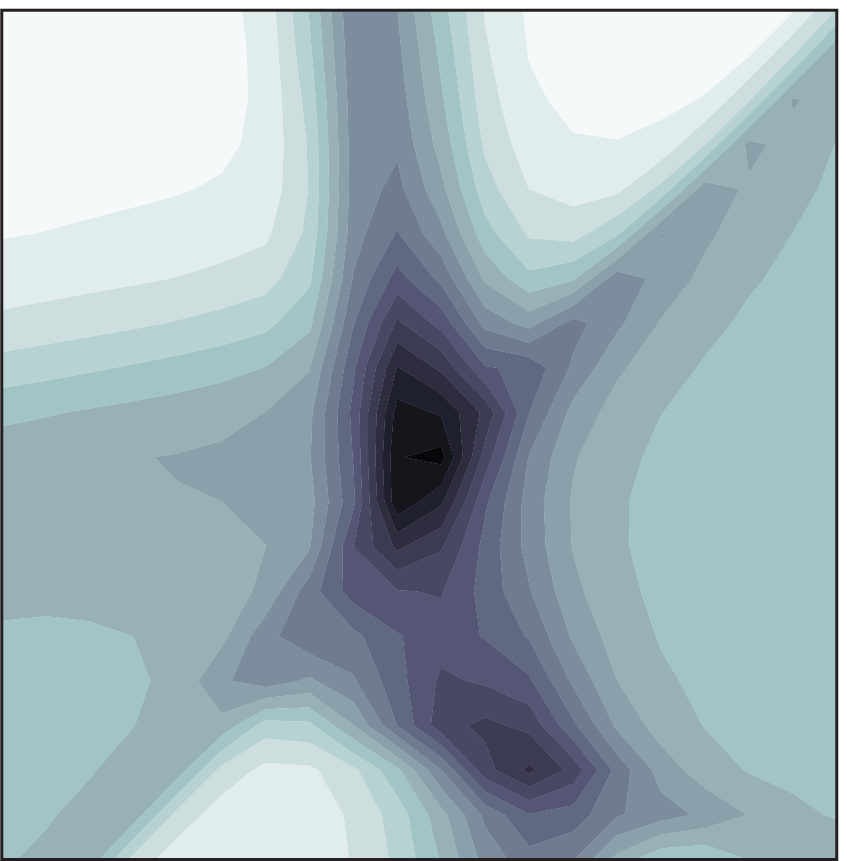}}
  \caption{In this toy example, we learn generative models for a 2D multivariate normal with identity covariance centered at (5, 5). (a) Explicit density models such as Normalizing Flows concentrate probability mass at the data distribution (b) Four independently trained GANs learn random discriminative boundaries, each corresponding to a different implied generator distribution. To ensure that the GAN discriminators form a clear discriminative boundary between $p(\rx)$ and $p_\theta(\rx)$, we train the discriminators an additional 10k steps to convergence. Each of these boundaries fails to enclose the true data distribution. (c) Predictive uncertainty over an ensemble of discriminators ``fences in'' the shared, low-variance region corresponding to $p(\rx)$.}
  \label{fig:toy2d}
\end{figure}

Figure~\ref{fig:toy2d}b illustrates a simple 2D density modeling task where individual GAN discriminators -- when trained to convergence -- learn a discriminative boundary that does not adequately capture $p(\rx)$. Unsurprisingly, a discriminative model tasked with classifying between $p(\rx)$ and $q(\rx)$ perform poorly when presented with inputs that belong to neither distribution. Despite this apparent shortcoming, GAN discriminators are still applied to successfully to anomaly detection problems \citep{schlegl2017unsupervised, deecke2018anomaly,kliger2018novelty}. 

Unlike discriminative anomaly classifiers, which model $p(\rx)/q(\rx)$ for a static $q(\rx)$, the generative distribution $p_\theta(\rx)$ of a GAN is trained jointly with the discriminator. The likelihood ratio $p(\rx)/p_\theta(\rx)$ learned by a GAN discriminator is uniquely randomized by GAN training dynamics on $\theta$ (Figure~\ref{fig:toy2d}b). By training an ensemble of GANs, we can recover an (unnormalized) approximation of $p(\rx)$ via decision boundaries between $p(\rx)$ and randomly sampled $p_\theta(\rx)$ (Figure~\ref{fig:toy2d}c). We implement an anomaly detector using the variance of the discriminator logit, and show that although ensembling GANs lead to far more effective OoD detection than a single GAN (Supplemental Material), it does not outperform our WAIC-based Generative Ensembles.

\section{Experimental Results}
\label{sec:experiments}

Following the experiments proposed by \cite{liang2017enhancing} and \cite{alemi2018uncertainty}, we train OoD models on MNIST, Fashion MNIST, CIFAR-10 datasets, and evaluate anomaly detection on test samples from other datasets. Source code for VAE and GAN experiments are located at \url{https://github.com/hschoi1/rich_latent}, and code for reproducing ODIN baselines are located at \url{https://github.com/ericjang/odin}.

We approximate likelihoods with two kinds of models, which we abbreviate as $p_\theta(\rx)$ in Table~\ref{table:results_table}. For MNIST and FashionMNIST, we estimate $\log p_\theta(\rx)$ using a 16-sample Importance-Weighted AutoEncoder (IWAE) bound, and for CIFAR-10, we compute exact densities via Glow \cite{kingma2018glow}. We follow the protocol as suggested by \cite{lakshminarayanan2017simple} to use 5 models with different parameter initializations, and train them independently on the full training set with no bootstrapping. For VAE architectures we also evaluate the rate term $\KL ( q_\theta(z|x) \Vert p(z) )$, which corresponds to information loss between the latent inference distribution and prior \cite{alemi2018uncertainty}. Glow uses a fixed isotropic Gaussian latent distribution, so we also devise a ``typicality-aware baseline'' that measures the euclidean distance $\normdiff$ to the closet point on a hypersphere of radius $\sqrt{3072}$. Finally, we report GAN-based anomaly detection in Appendix \ref{appendix:gan}.
We extend the VAE example code\footnote{\url{https://github.com/tensorflow/probability/blob/6dfeb4ba7a5164fe4617e4967b168640109221c9/tensorflow_probability/examples/vae.py}} from Tensorflow Probability~\citep{dillon2017tensorflow} to use a Masked Autoregressive Flow prior \citep{papamakarios2017masked}, and train the model for 5k steps: The encoder consists of 5 convoluton layers. The decoder consists of 6 deconvolution layers followed by a convolution layer, and is trained using maximum likelihood under independent Bernoullis. We use a flexible learned prior $p_\theta(\rz)$ in our VAE experiments, but did not observe a significant performance difference compared to the default mixture prior in the base VAE code sample. We use an alternating chain of 6 MAF bijectors and 6 random permutation bijectors. Models are trained with Adam ($\alpha=1\mathrm{e}{-3}$) with cosine decay on learning rate. Each Glow model uses the default implementation and training parameters from the Tensor2Tensor project \cite{tensor2tensor} and is trained for 200k steps on a single GPU.

The baseline methods, ODIN and VIB, are dependent on a classification task and learn from the joint distribution of images and labels, while our methods use only images. For the VIB baseline, we use the rate term as the threshold variable. The experiments in \cite{alemi2018uncertainty} make use of (28, 28, 5) ``location-aware'' features concatenated to the model inputs, to assist in distinguishing spatial inhomogeneities in the data. In this work we train vanilla generative models with no special modifications, so we also train VIB without location-aware features. For CIFAR-10 experiments, we train VIB for 26 epochs and converge at 75.7\% classification accuracy on the test set. All other experimental parameters for VIB are identical to those in \cite{alemi2018uncertainty}.

\begin{table}[t]
  \caption{We train models on MNIST, Fashion MNIST, and CIFAR-10 and compare OoD classification ability to baseline methods using the threshold-independent Area Under ROC curve metric (AUROC). $p_\theta(\rx)$ is a single-model likelihood approximation (IWAE for VAE, log-density for Glow). WAIC is the Watanabe-Akaike Information Criterion as estimated by the Generative Ensemble. ODIN results reproduced from \cite{liang2017enhancing}. Best results for each task shown in bold.}
  \label{table:results_table}
  
\vskip 0.15in
\begin{center}
\begin{footnotesize}
\begin{sc}
  \begin{tabular}{lccccl}
    \toprule
 OoD & ODIN & VIB & Rate & $p_\theta(\rx)$ & WAIC  \\
    \midrule 
    MNIST & VAE \\
    \midrule 
  Omniglot & \textbf{100} & 97.1 & 99.1 & 97.9 & 98.5 \\
      notMNIST & 98.2 & 98.6 & 99.9 & \textbf{100} & \textbf{100} \\
      FashionMNIST & N/A & 85.0 & \textbf{100} & \textbf{100} & \textbf{100} \\
      Uniform & \textbf{100} & 76.6 & \textbf{100} & \textbf{100} & \textbf{100} \\ 
      Gaussian & \textbf{100} & 99.2 & \textbf{100} & \textbf{100} & \textbf{100} \\
      HFlip & N/A & 63.7 & 60.0 & 84.1 & \textbf{85.0} \\
      VFlip & N/A & 75.1 & 61.8 & 80.0 & \textbf{81.3} \\
      Adv & N/A & N/A & \textbf{100} &  0 & \textbf{100} \\
    \midrule 
    FashionMNIST & VAE\\
    \midrule 
 Omniglot & N/A & \textbf{94.3} & 83.2 & 56.8 & 79.6 \\
           notMNIST & N/A & 89.6 & 92.8 & 92.0 & \textbf{98.7} \\
     MNIST & N/A & \textbf{94.1} & 87.1 & 42.3 & 76.6 \\
               Uniform & N/A & 79.6 & 99.0 & \textbf{100} & \textbf{100} \\
               Gaussian & N/A & 89.3 & \textbf{100} & \textbf{100} & \textbf{100} \\
               HFlip & N/A & \textbf{66.7} & 53.4 & 59.4 & 62.4 \\
               VFlip & N/A & \textbf{90.2} & 58.6 & 66.8 & 74.0 \\
               Adv & N/A & N/A & \textbf{100} & 0.1 & \textbf{100} \\
    \midrule 
     OoD & ODIN & VIB & $\normdiff$ & $p_\theta(\rx)$ & WAIC \\               
    \midrule
    CIFAR-10 & Glow \\
    \midrule
    CelebA & 85.7 & 73.5 & 22.9 & 75.6 & \textbf{99.7}  \\
          SVHN & 89.9 & 52.8  & 74.4 & 7.5 & \textbf{100} \\
          ImageNet32 & 98.5 & 70.1 & 12.3 & 93.8 & \textbf{95.6}  \\
          Uniform & 99.9 & 54.0 & \textbf{100} &  \textbf{100} & \textbf{100}  \\ 
          Gaussian &\textbf{100} & 45.8 & \textbf{100} &  \textbf{100} & \textbf{100}  \\
          HFlip & 50.1 & \textbf{50.6} & 46.2 & 50.1 & 50.0 \\
          VFlip & 84.2 & \textbf{51.2} & 44.0 & 50.6 & 50.4  \\     
    \bottomrule
 \end{tabular}
 
\end{sc}
\end{footnotesize}
\end{center}
\vskip -0.1in
 
\end{table}

Despite being trained without labels, our methods -- in particular Generative Ensembles -- outperform ODIN and VIB on most OoD tasks. The VAE rate term is robust to adversarial inputs on the IWAE, because the FGSM perturbation primarily minimizes the (larger) distortion component of the variational lower bound. The performance of VAE rate versus VIB also suggests that latent codes learned from generative objectives are more useful for OoD detection that latent codes learned via a classification-specific objective. Surprisingly, despite not being a general estimate of typicality, WAIC dramatically outperforms the $\normdiff$ baseline.

\begin{figure}[h]
  \centering
  \includegraphics[width=\linewidth]{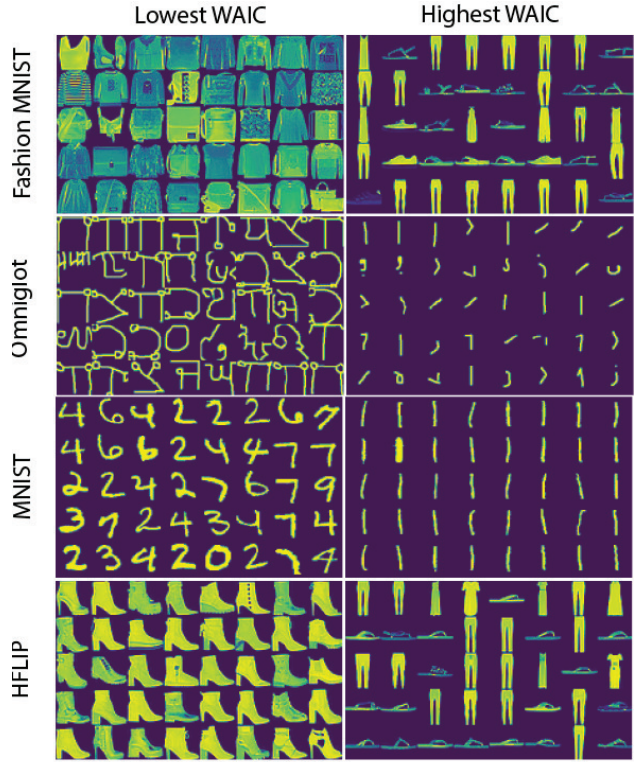} 
  \caption{Left: lowest WAIC for each evaluation dataset. Right: highest WAIC for each evaluation dataset.}
  \label{fig:lowest-highest}
\end{figure}

We present in Figure~\ref{fig:lowest-highest} images from OoD datasets with the smallest and largest WAICs, respectively. Models trained on FashionMNIST tend to assign higher likelihoods to straight, vertical objects like ``1's'' and pants. It is not altogether surprising that AUROC scores on on the HFlip transformation was low, given that many pants, dresses are symmetric.

\subsection{Credit Card Anomaly Detection}

We consider the problem of detecting fraudulent credit card transactions from the Kaggle Credit Fraud Challenge \citep{dal2015calibrating}. A conventional approach to fraud detection is to include a small fraction of fraudulent transactions in the training set, and then learn a discriminative classifier. Instead, we treat fraud detection as an anomaly detection problem where only normal credit card transactions are available at training time. This is motivated by realistic test scenarios, where an adversary is hardly restricted to generating data identically distributed to the training set.

We compare single likelihood models (16-sample IWAE) and Generative Ensembles (ensemble variance of IWAE) to a binary classifier baseline that has access to a training set of fraudulent transactions in Table~\ref{table:creditcard}. The classifier baseline is a fully-connected network with 2 hidden ReLU layers of 512 units, and is trained using a weighted sigmoid cross entropy loss (positive weight=580) with Dropout and RMSProp ($\alpha=1\mathrm{e}{-5}$). The VAE encoder and decoder are fully connected networks with single hidden layers (32 and 30 units, respectively) and trained using Adam ($\alpha=1\mathrm{e}{-3}$).

Unsurprisingly, the classifier baseline performs best because fraudulent test samples are distributed identically to fraudulent training samples. Even so, the single-model density estimation and Generative Ensemble achieve reasonable results.

\subsection{Failure Analysis}

In this section we discuss the experiments in which Generative Ensembles performed poorly, and suggest simple fixes to address these issues.

\begin{figure*}[t]
  \centering
  \includegraphics[width=\textwidth]{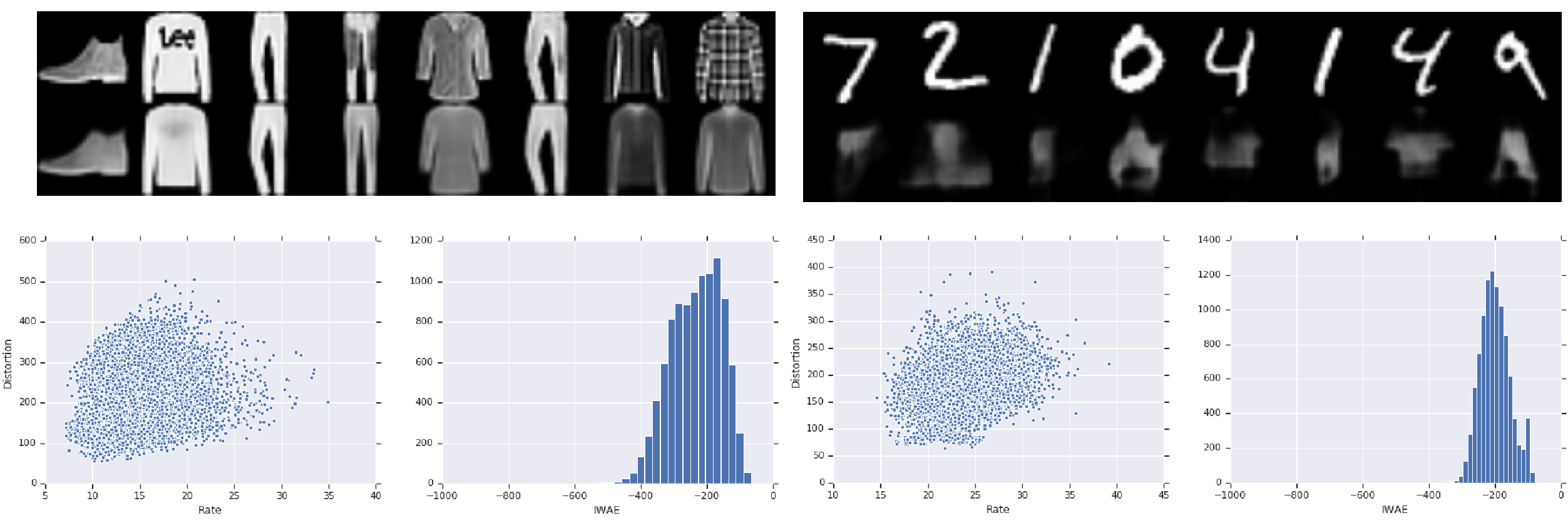}
  \caption{Top: Inputs and decoded outputs from a VAE trained on Fashion MNIST($\beta=1$) for Fashion MNIST (left) and MNIST (right). Although Fashion MNIST inputs appear to be better reconstructed (suggesting higher likelihoods), they have comparable distortions to MNIST. The bottom row shows that Fashion MNIST and MNIST test samples have comparable rate-distortion scatter plots and IWAE histograms. This results in poor OoD detection of MNIST on Fashion MNIST images, which is mitigated by training with $\beta=.1$ to encourage better auto-encoding.}
  \label{fig:fmnist_vae_decode}
\end{figure*}

In our early experiments, we found that a VAE trained on Fashion MNIST performed poorly on all OoD datasets when using $p_\theta(\rx)$ and WAIC metrics. This was surprising, since the same metrics performed well when the same VAE architecture was trained on MNIST. To explain this phenomenon, we show in Figure~\ref{fig:fmnist_vae_decode} inputs and VAE-decoded outputs from Fashion MNIST and MNIST test sets. Fashion MNIST images are reconstructed properly, while MNIST images are are barely recognizable after decoding.

A VAE’s training objective can be interpreted as the sum of a pixel-wise autoencoding loss (distortion) and a ``semantic'' loss (rate). Even though Fashion MNIST appears to be better reconstructed in a semantic sense, the distortion values between the FashionMNIST and MNIST test datasets are numerically quite similar, as shown in Figure~\ref{fig:fmnist_vae_decode}. Distortion terms make up the bulk of the IWAE predictions in our models, thus explaining why $p_\theta(\rx)$ was not very discriminative when classifying OoD MNIST examples.

\cite{higgins2016beta} propose $\beta$-VAE, a simple modification to the standard VAE objective: $p(x | z) + \beta \cdot \KL ( q_\theta(z|x) \Vert p(z) )$. $\beta$ controls the relative balance between rate and distortion terms during training. Setting $\beta < 1$ is a commonly prescribed fix for encouraging VAEs to approach the ``autoencoding limit'' and avoid posterior collapse \citep{alemi2018fixing}. At test time, this results in higher-fidelity autoencoding at the expense of higher rates, which seems to be a more useful signal for identifying outliers than the total pixel distortion (also suggested by Table~\ref{table:results_table}, column 4). Re-training the ensemble with $\beta = .1$ encourages a higher distortion penalty during training, and thereby fixes the OoD detection model.


\begin{table}[h]
 \caption{Comparison of density-based anomaly detection approaches to a classification baseline on the Kaggle Credit Card Fraud Dataset. The test set consists of 492 fraudulent transactions and 492 normal transactions. Threshold-independent metrics include False Positives at 95\% True Positives (FPR@95\%TPR), Area Under ROC (AUROC), and Average Precision (AP). Density-based models (Single IWAE, WAIC) are trained only on normal credit card transactions, while the classifier is trained on normal and fraudulent transactions. Arrows denote the direction of better scores.}
\label{table:creditcard}
\vskip 0.15in
\begin{center}
\begin{small}
\begin{sc}

\begin{tabular}{llll}
\toprule
Method    &FPR@95\%TPR  $\downarrow$ &AUROC   $\uparrow$  &AP $\uparrow$ \\
\midrule
Classifier &4.0 & 99.1 & 99.3 \\
Single IWAE & 15.7 & 94.6 & 92.0 \\
WAIC & 15.2 & 94.7 & 92.1 \\
    \bottomrule
 \end{tabular}

\end{sc}
\end{small}
\end{center}
\vskip -0.1in
\end{table}

\section{Discussion and Future Work}
Out-of-Distribution (OoD) detection is a critical piece of infrastructure for ML applications where the test data distribution is not known at training time. Our paper has two main contributions. The more straightforward one is a novel technique for OoD detection in which ensembles of generative models can be used to estimate WAIC. We perform a comparison to prior techniques published in OoD literature to show that it is competitive with past models and argue for why OoD detection should be done “model-free”.

The other contribution is the intriguing observation that WAIC shouldn’t work! It has been established in the literature~\cite{nalisnick2018dodeep, hendrycks2018deep} that likelihood alone is not sufficient for determining whether data is out of distribution. Simply put, regions with high likelihood are not necessarily regions with high probability mass. It is premature, however, to invoke this explanation as the sole reason generative models trained on CIFAR-10 fail to identify SVHN as OoD, as we show, \emph{e pur si muove}, that robust likelihood measures like WAIC \emph{can} distinguish the samples correctly. As we show in Figure~\ref{fig:radii_interior}, the maximum likelihood objective typically used in flow-based models forces the training distribution (CIFAR-10) towards the region of maximum likelihood in latent space. We hypothesize that these datasets are quite different, different enough that the SVHN images likelihoods can be very sensitive to the initial conditions and architectural hyperparameters of the trained generative models. The surprising effectiveness of WAIC motivates further investigation into robust measures of typicality which incorporate both likelihood as well as some notion of local volume, as it is clear that there is an incomplete theoretical understanding within the research community of the limits of likelihood based OOD detection.

Furthermore, the observation that CIFAR-10 is mapped to an annulus of radius $< \sqrt{3072}$ is in itself quite disturbing, as it suggests that better flow-based generative models (for sampling) can be obtained by encouraging the training distribution to overlap better with the typical set in latent space.

\section*{Acknowledgements}
We thank Manoj Kumar, Peter Liu, Jie Ren, Justin Gilmer, Ben Poole, Augustus Odena, and Balaji Lakshminarayanan for code and valuable discussion. We would also like to thank all the organizers and participants of DL Jeju Camp 2018.


\bibliography{icml2019_arxiv}
\bibliographystyle{icml2019}
\vfill\null  

\pagebreak
\onecolumn
\appendix

\section{Terminology and Abbreviations}

\begin{tabular}{p{1.25in}p{3.25in}}
$p(\rx)$ & Training data distribution \\
$q(\rx)$ & OoD data distribution \\
$p_\theta(\rx)$ & Learned approximation of true distribution with parameters $\theta$. May be implicitly specified, i.e. a fully-observed density model \\
$q_\theta(\rx)$ & Learned approximation of OoD distribution with parameters $\theta$. May be implicitly specified, i.e. via a GAN discriminator that learns $p(x)/q_\theta(\rx)$ \\
\midrule
OoD Input & Out-of-Distribution Input. Invalid input to a ML model \\
Anomaly & Synonym with OoD Input \\
Epistemic Uncertainty & Variance in a model's predictive distribution arising from ignorance of true model parameters for a given input \\
Aleatoric Uncertainty & Variance in a model's predictive distribution arising from inherent, irreducible noise in the inputs \\
Predictive Uncertainty & Variance of a model's predictive distribution, which takes into account all of the above \\
\midrule
MNIST & Dataset of handwritten digits (size: 28x28) \\
FashionMNIST & Dataset of clothing thumbnails (size: 28x28) \\
CIFAR-10 & Dataset of color images (size: 32x32x3) \\
\midrule
GAN & Generative Adversarial Network. See \cite{goodfellow2014generative} \\
FSGM & Fast Sign Gradient Method \\
WGAN & Wasserstein GAN. See \cite{arjovsky2017wasserstein} \\
VAE & Variational Autoencoder. See \cite{kingma2013auto, rezende2014stochastic} \\
Rate & $\KL ( q_\theta(z|x) \Vert p(z) )$ term in the VAE objective. Information loss between encoder distribution and prior over latent code \\
IWAE & Importance Weighted Autoencoder \\
GLOW & A generative model based on normalizing flows. See \cite{kingma2018glow} \\
\midrule
ODIN & Out-of-DIstribution detector for Neural networks. See \cite{liang2017enhancing} \\
VIB & Variational Information Bottleneck. See \cite{alemi2018uncertainty} \\
WAIC & Watanabe-Akaike Information Criterion. See \cite{watanabe2010asymptotic} \\
\midrule
AUROC & Area Under ROC Curve \\
FPR@95\%TPR & False Positives at 95\% True Positives \\
AP & Average Precision \\
\end{tabular}

\section{OoD Detection with GAN Discriminators}
\label{appendix:gan}
For MNIST, Fashion MNIST, and CIFAR-10 datasets, we train WGANs and fine tune the discriminators on stationary $p(\rx)$ and $p_\theta(\rx)$. Table~\ref{table:supplemental_results_table} shows that ensemble variance of the discriminators, $\Var(D)$. Our WGAN model's generator and discriminator share the same architecture with the VAE decoder and encoder, respectively. The discriminator has an additional linear projection layer to its prediction of the Wasserstein metric. We train all WGAN models for 20k generator updates. To ensure $D$ represents a meaningful discriminative boundary between the two distributions, we freeze the generator and fine-tune the discriminator for an additional 4k steps on stationary $p(\rx)$ and $p_\theta(\rx)$. For CIFAR-10 WGAN experiments, we change the first filter size in the discriminator from 7 to 8.

\begin{table}[h]
  \caption{We train models on MNIST, Fashion MNIST, and CIFAR-10 and compare OoD classification ability to one another using the threshold-independent Area Under ROC curve metric (AUROC). $D$ corresponds to single WGAN discriminators with 4k fine-tuning steps on stationary $p(\rx)$, $q_\theta(\rx)$. $\Var(D)$ is uncertainty estimated by an ensemble of discriminators. Rate is the $\KL$ term in the VAE objective. Best results for each task shown in bold.}
  \label{table:supplemental_results_table}
  
\vskip 0.15in
\begin{center}
\begin{small}
\begin{sc}
  \begin{tabular}{lccc}
    \toprule
  OoD & D & Var(D)  \\
    \midrule 
    MNIST \\
    \midrule 
        Omniglot  & 52.7 & 81.2  \\
        notMNIST  & 92.7 & 99.8 \\
        Fashion MNIST  & 81.5 & \textbf{100} \\
        Uniform  & 93.9 & \textbf{100}  \\ 
        Gaussian  & 0.8 & \textbf{100}  \\
        HFlip & 44.5 & 60.0 \\
        VFlip & 46.1 & 61.8 \\
        Adv & 0.7 & \textbf{100} \\
    \midrule 
    FashionMNIST \\
    \midrule 
       Omniglot  & 19.1 & 83.2 \\
       notMNIST  & 17.9 & 92.8   \\
       MNIST  & 45.1 & 87.1  \\
       Uniform  & 0 & \textbf{99.0} \\
       Gaussian & 0 & \textbf{100} \\
       HFlip & 54.7 & 53.4  \\
       VFlip  & 67.2 & 58.6  \\
       Adv  & 0 & \textbf{100} \\
    \midrule 
    CIFAR-10 \\
    \midrule 
           CelebA  & 57.2 & \textbf{74.4} \\
           SVHN  & \textbf{68.3} & 62.3  \\
           ImageNet32  & 49.1 & \textbf{62.6} \\
           Uniform  & \textbf{100} & \textbf{100}   \\ 
           Gaussian  & \textbf{100} & \textbf{100}   \\
           HFlip  & \textbf{51.6} & 51.3  \\
           VFlip  & \textbf{59.2} & 52.8  \\
           
    \bottomrule
 \end{tabular}
\end{sc}
\end{small}
\end{center}
\vskip -0.1in
 
\end{table}

\section{VAE Architectural Details}
\label{appendix:maf_vae}

We use a flexible learned prior $p_\theta(\rz)$ in our VAE experiments, but did not observe a significant performance difference compared to the default mixture prior in the base VAE code sample. We use an alternating chain of 6 MAF bijectors and 6 random permutation bijectors. Each MAF bijector uses TensorFlow Probability's default implementation with the following parameter: 

\begin{lstlisting}
shift_and_log_scale_fn=tfb.masked_autoregressive_default_template(
  hidden_layers=(512, 512))
\end{lstlisting}

Models are trained with Adam ($\alpha=1\mathrm{e}{-3}$) with cosine decay on learning rate.


\end{document}